\documentclass{article}
\usepackage{spconf,amsmath,graphicx,adjustbox}
\usepackage{lmodern}
\usepackage[T1]{fontenc}

\title{Enhancing the Prediction of Emotional Experience in Movies using Deep Neural Networks: The Significance of Audio and Language }

\name{Sogand Mehrpour Mohammadi$^1$, Meysam Gouran Orimi$^2$, Hamidreza Rabiee$^3$} 
\address{$^1$Department of Computer Science, University of Mazandaran, Babolsar, Iran \\
$^2$Department of Computer Engineering, Shahrood University of Technology, Shahrood, Iran\\
$^3$Department of Electrical Engineering, Islamic Azad University, Karaj Branch, Karaj, Iran \\Email: hr.rabiee@iau.ac.ir\\}

\begin{document}
%
\maketitle
\begin{abstract}
Our paper focuses on making use of deep neural network models to accurately predict the range of human emotions experienced during watching movies. In this certain setup, there exist three clear-cut input modalities that considerably influence the experienced emotions: visual cues derived from RGB video frames, auditory components encompassing sounds, speech, and music, and linguistic elements encompassing actors' dialogues.
Emotions are commonly described using a two-factor model including valence (ranging from happy to sad) and arousal (indicating the intensity of the emotion).
In this regard, a Plethora of works have presented a multitude of models aiming to predict valence and arousal from video content.
However, non of these models contain all three modalities, with language being consistently eliminated across all of them.
 In this study, we comprehensively combine all modalities and conduct an analysis to ascertain the importance of each in predicting valence and arousal.  
 Making use of pre-trained neural networks, we represent each input modality in our study. In order to process visual input, we employ pre-trained convolutional neural networks to recognize scenes~\cite{zhou2016places}, objects~\cite{szegedy2017inception}, and actions~\cite{carreira2017quo,8354292}. For audio processing, we utilize a specialized neural network designed for handling sound-related tasks, namely SoundNet ~\cite{aytar2016soundnet}. Finally, Bidirectional Encoder Representations from Transformers (BERT) models are used to extract linguistic features~\cite{devlin2018bert} in our analysis. We report results on the COGNIMUSE dataset~\cite{zlatintsi2017cognimuse}, where our proposed model outperforms the current state-of-the-art approaches. Surprisingly, our findings reveal that language significantly influences the experienced arousal, while sound emerges as the primary determinant for predicting valence. In contrast, the visual modality exhibits the least impact among all modalities in predicting emotions.
\end{abstract}
\begin {keywords}
Emotion Recognition, Multimodal, Sound, Text, Spatio-temporal, Affective Computing
\end{keywords}
\section{Introduction}

In recent years, notable advancements have been made in the field of emotion recognition specifically pertaing to video and sound analysis~\cite{zhou2016places}. However, there remain challenges as predictive models need to improve their accuracy and information processing capabilities to efficiently disentangle the influence of diverse modalities. 
These advancements will empower multimedia creators in the advertising or film industries to employ these improved predictive models as valuable tools. Researchers in the fields of neuroscience and psychology have extensively investigated the impacts of various input stimulus modalities on evoked emotions~\cite{schmidt2001frontal,sammler2007music,article} and ~\cite{koelsch2010towards,zentner2008emotions}. Multimodal approaches to emotion identification are considered among the most important techniques for obtaining computer intelligence, as humans can perceive and comprehend emotions via the combined information conveyed in sound, video, and text~\cite{huang2019speech,huang2019attention}.
In the past, research on emotion recognition has predominantly  directed on discriminative emotion features and recognition models that rely on single modalities such as audio signals~\cite{hu2017learning} or facial expressions
in video. In this work, we place significant emphasis on considering context information, knowing its importance in developing a robust emotion prediction model~\cite{lugovic2016techniques}. There is a scarcity of research focused on anticipating the feeling responses that videos and other multimedia content can elicit in viewers. With the progression of  neural networks, numerous studies have endeavored to extract suitable global and local features from raw speech signals and RGB images, leading to improved performance in emotion recognition tasks~\cite{zhu2017emotion}. These collective efforts have enhanced the performance of emotion recognition systems. Recent advancements in emotion recognition systems are hindered by a lack of comprehensive multimodal analysis, as they fail to consider all the relevant modalities that influence the recognition of emotions.

\begin{figure}[!t]
\begin{center}
   \includegraphics[width=1\linewidth]{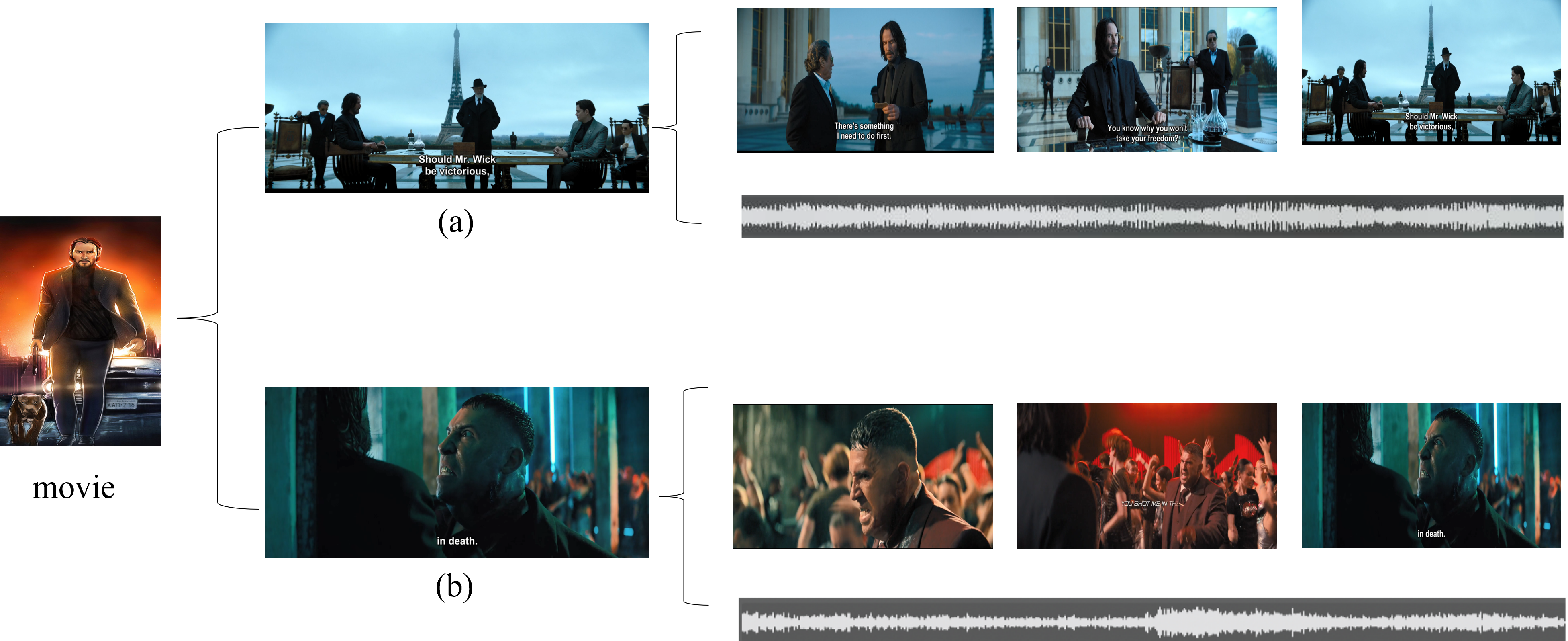}
\end{center}
   \caption{ Sample frames from the COGNIMUSE dataset to show our different modalities, i.e., video frame, sound and subtitles,(a) low valence, (b) high valence.}
\label{fig:sample}
\end{figure}

\begin{figure*}[t]
\begin{center}
   \includegraphics[width=.9\linewidth]{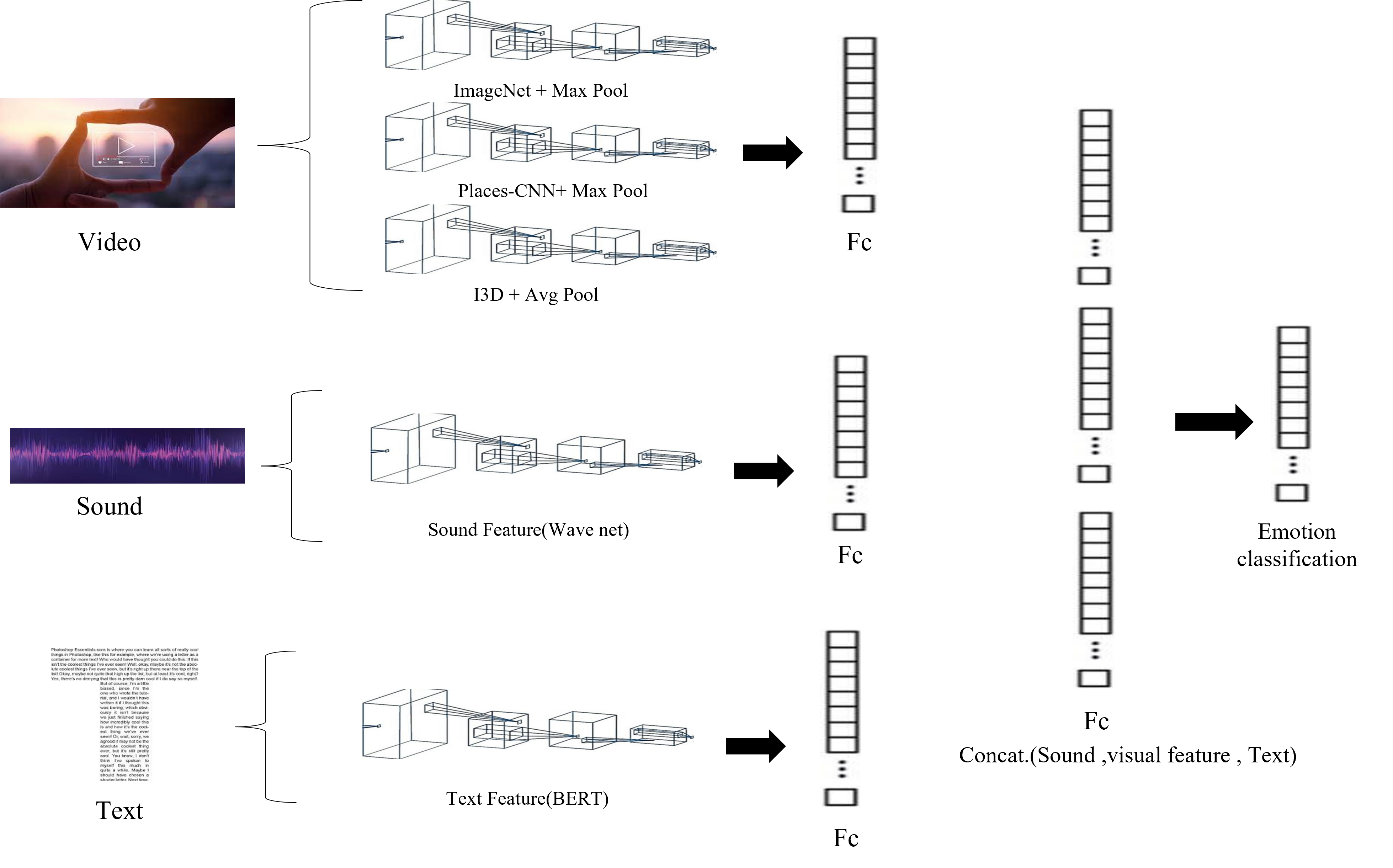}
\end{center}
   \caption{Video, Sound, and text modalities pass through our proposed network for emotion classification using fully connected layers.}
\label{fig:network}
\end{figure*}

In this study, we harness the capabilities of deep convolutional neural networks (CNNs) to present a novel three-dimensional model of effective content. This model enables us to precisely predict the evoked emotion from movies, music, and text, with the ability to independently forecast arousal and valence as distinct dimensions of emotion.
To extract features from static RGB frames, we employ state-of-the-art deep convolutional neural networks (CNNs). These advanced CNN models empower us to capture both picture-related features and motion-related features from the RGB frames.
The I3D~\cite{carreira2017quo} architecture, which combines spatial and temporal networks, is employed to determine activity in a manner similar to CNNs. In order to extract information from scenes~\cite{zhou2016places} and objects~\cite{szegedy2017inception} in video frames, the spatial network is used. On the other hand, the temporal network captures and learns information according to the motion of the camera and objects across successive frames.
Furthermore, sound representation is computed using soundNet~\cite{aytar2016soundnet}, which is employed to extract the audio features. SoundNet has illustrated notable performance in the classification of acoustic scenes and objects by utilizing the inherent synchronization between vision and sound during model training. This natural synchronization allows for more efficient utilization of the advantages presented by both visual and audio information. To obtain text features from movie subtitles, we employed the pre-trained Bidirectional Encoder Representations from Transformers (BERT) model~\cite{devlin2018bert}. By feeding the movie subtitles into the BERT model, we achieved meaningful text representations that are vital for emotion recognition. In other words, the BERT model is utilized for extracting deep bidirectional representations from unlabeled text. In our study, we employed fully connected layers to integrate proposed multimodal networks. This integration was performed using the expanded COGNIMUSE dataset, ~\cite{zlatintsi2017cognimuse}.

\subsection{Related work}
Currently, the majority of computer vision research on emotion identification mainly focuses on analyzing human facial expressions in both video and still images~\cite{cohn2014automated}. When inspecting research related to predicting emotions from videos, it is evident that the majority of works utilize multimodal techniques that consider and combine data from diverse modalities. Moreover, the field of computer vision has shown a growing interest in exploring the relationships and interactions between diverse modalities, including the synergies between sound and vision modalities ~\cite{song2012multimodal}.
In this paper, we aim to explore the relationship between sound and visual semantic information in the domain of emotion detection. While the combination of diverse modalities, such as sound and language, has been increasingly studied in the field of speech recognition ~\cite{slaney2002semantic}, we extend this exploration to the realm of emotion detection, searching for uncover how these modalities synergistically contribute to the exact determination and understanding of emotions in multimedia content.
Additionally, the relationship between text and visual modalities has been thoroughly investigated for the question-answering task ~\cite{anderson2018bottom}.In this paper, our main goal is to develop a comprehensive representation that empowers us to conceive emotions through the integration of multiple modalities, including text, sound, and visual information extracted from both static images and motion. Unlike previous works that concentrates on only two modalities, our aim is to capture the rich and complex nature of emotions by harnessing the synergistic interaction of all available modalities. To extract text features from the movie subtitles in our dataset, we utilized a pre-trained BERT model~\cite{devlin2018bert}.By feeding the movie subtitles into this pre-trained BERT model, we can extract meaningful text features that capture the linguistic content and context for further analysis and emotion anticipation. 
The main contribution of our paper is to demonstrate the effectiveness of a deep model that leverages three important natural modalities for emotion recognition. We provide an in-depth description of our methodology and thorough explanations of the conducted experiments in the following sections. Through these detailed analyses, we aim to showcase the robustness and accuracy of our approach in recognizing and predicting emotions across several modalities. In Section 2, we present our approach for developing a comprehensive understanding of emotion representations by incorporating concepts from sound, text, and visuals. In section 3, we present a detailed account of the experiments conducted to determine the effectiveness of our proposed representations. We present various experimental setups and analyze the performance of our model in anticipating and classifying emotions based on the extracted features. The experimental results serve to validate the robustness and efficacy of our approach across diverse datasets and scenarios.

\setlength{\tabcolsep}{0.1pt}
\begin{table*}[t!]
\begin{center}
\scalebox{.08}{}
\begin{tabular}{|c|c|c|c|c|}
\hline
\multicolumn{1}{|l|}{\textbf{}}&\multicolumn{2}{||c||}{\textbf{Arousal}} & \multicolumn{2}{c|}{\textbf{valence}} \\ \hline
\hline 
       \textbf{model} & \textbf{Accuracy(\%)} & \textbf{Accuracy} $\pm$ \textbf{1 (\%)} & $\;\;$ \textbf{Accuracy (\%)}& \textbf{Accuracy} $\pm$ \textbf{1 (\%)}  \\  \hline 
       \textbf{FC (RGB frame + OF + Audio)}~\cite{thao2019multimodal} &53.32&        94.75&        43.10&       90.51 \\ \hline
       \textbf{LSTM (RGB frame + OF + Audio)}~\cite{thao2019multimodal} &48.64 &               95.28 &           37.20&        89.22 \\ \hline
\hline
      \textbf{Visual (Resnet + Places + I3D)} &      42.23 &      90.55 &           41.71 &        91.23 \\ \hline
      \textbf{Sound (SoundNet)} &       58.59 &       95.11&  \textbf{56.28} &       \textbf{97.20} \\ \hline
      \textbf{Text (BERT)} &       \textbf{58.86} &       \textbf{95.13}&     32.55  &       82.62 \\ \hline
      \textbf{Visual+ Sound} &   54.30    &   94.20    &  32.30       &   82.90    \\ \hline
      \textbf{Resnet+ Sound} &   58.59  & 95.11   &      36.43 &     86.20  \\ \hline
      \textbf{Text+ Sound} &     58.59 &    95.11   &     30.82    &      83.17 \\ \hline
      \textbf{Text + visual} &    54.22   &  94.36     &   30.40         &  83.22      \\ \hline
      \textbf{Text + Resnet} &   58.22    &  95.11     &     36.32       &  86.18     \\ \hline
      
      \textbf{Visual+ Sound + Text}&  56.86 &      94.65 & 42.19    &       91.46 \\ \hline
      
\end{tabular}
\end{center}
\caption{Accuracy results on COGNIMUSE dataset along with the arousal and valence experienced emotion labels. }
\label{table:results101}
\end{table*}
\setlength{\tabcolsep}{0.pt}
\setlength{\tabcolsep}{0.1pt}
\begin{table*}[t!]
\begin{center}
\scalebox{0.8}{}
\begin{tabular}{|c||c|c||c|c|}
\hline
\multicolumn{1}{|l|}{\textbf{}}&\multicolumn{2}{||c||}{\textbf{Arousal}} & \multicolumn{2}{c|}{\textbf{valence}} \\ \hline
\hline
       \textbf{Model} & \textbf{Accuracy (\%)} & \textbf{Accuracy} $\pm$ \textbf{1 (\%)} & $\;\;$ \textbf{Accuracy (\%)}& \textbf{Accuracy} $\pm$ \textbf{1 (\%)}     \\  \hline
       \textbf{FC (RGB frame + OF + Audio)}~\cite{thao2019multimodal} &31.20&        72.94&        30.33&       66.95 \\ \hline
       \textbf{LSTM (RGB frame + OF + Audio)}~\cite{thao2019multimodal} &       30.80 &      71.69 &     22.54&     57.63 \\ \hline
       \textbf{Malandrakis et al.}~\cite{malandrakis2011supervised} &     24.00&      57.00 &          24.00 &       64.00 \\ \hline
\hline
      \textbf{Visual (Resnet + Places + I3D)} &      42.67 &    86.99 & 49.20 &       93.81  \\ \hline
      \textbf{Sound (SoundNet)} &  58.51  &    95.10   &  \textbf{55.85}   &     96.21   \\ \hline
      \textbf{Text (BERT)} &     \textbf{58.56}   &      95.10 &   32.45  &     83.99    \\ \hline
      \textbf{Visual+ Sound} &     58.51  &    95.10   &       54.85  &    \textbf{ 97.45}    \\ \hline
      \textbf{Text+ Sound} &   58.51  &    95.10     &     41.79     &    83.65    \\ \hline
      \textbf{Text + visual} &      58.51 &     94.94  &         31.53    &    84.21    \\ \hline
      \textbf{Text + Resnet} &     56.89   &    94.79 &        30.41    &    83.70   \\ \hline
      \textbf{Sound + Text + Resnet} &    54.42   &   94.07    &         37.48  &      87.23      \\ \hline
      \textbf{Visual+ Sound + Text}&   54.45 &      \textbf{96.35} &    32.48  &      83.99 \\ \hline
\end{tabular}
\end{center}
\caption{ Accuracy results on COGNIMUSE dataset along with the arousal and valence dimension on intended emotion label.}
\label{table:results102}
\end{table*}
\setlength{\tabcolsep}{0.pt}

\section{Multimodal Learning of Concepts}
In this section, our main goal is to achieve a comprehensive understanding of a given notion by exploring and comprehending its various modalities, including text, sound, and still images. By considering and integrating information from several modalities, we aim to capture a more complete and nuanced understanding of the targeted notion. We introduce a novel multimodal approach that utilizes all available modalities, including visual, sound, and text characteristics, to train our emotion prediction model. To achieve this, we employ the visual modality, which encompasses  scene comprehension, object detection, and motion data. Additionally, the acoustic content of the second modality is processed. Last but not least, we employ text processing techniques to train multimodal networks within the context of textual information. To begin with, our network is organized to automatically detect visual concepts in both spatial and temporal dimensions, addressing the numerous challenges according to this task.
Secondly, in order to tackle the sound processing challenge, it is vital for us to comprehend the audio segment that corresponds to a video concept.
Finally, we extract the wording directly from the movie's subtitles, which may correspond to specific scenes, several instances, or different visual ideas. We utilize a pre-trained network to learn representations range from sound, vision, and text modalities in order to recognize emotions. Prior to combining them to produce audio-visual-text features, each of these input categories is passed through fully connected layers to decrease dimensionality and adapt the representations for emotion prediction.
During the training phase of our presented network architecture, the weights of these fully connected layers are learned. Figure \ref{fig:sample} shows a few frame samples showing strong and low-valence emotions.

\subsection{Visual concept}
\textbf{\textit{Still Image:}} Following a similar approach to previous work~\cite{thao2019multimodal}, we divided the videos into five-second segments and extracted frames from each segment. Then, to extract features from static RGB frames, we employ the pre-trained Resnet152~\cite{szegedy2017inception} algorithm. The Imagenet dataset, which offers spatial information on the category of objects, is used to train the Resnet network.
\newline\textbf{\textit{Scene information:}} To ascertain the semantic importance of visual attributes in scenes, we utilize pre-trained VGG networks ~\cite{simonyan2014very} that were trained on Places datasets ~\cite{zhou2016places} for the scene classification task.To feed the fully connected layer of our network, we calculated the visual features from successive frames for each movie. After getting the visual features, we utilized max-pooling to downsample the feature representation and decrease its dimensionality.
\newline\textbf{\textit{Motion information:}} To retrieve motion characteristics from the input image frame sequence, we utilized a two-stream inflated 3D ConvNet (I3D) ~\cite{carreira2017quo}. The I3D model includes 2D ConvNet inflation and was trained on the Kinetics dataset for the action detection task.To extract spatio-temporal features, we applied a two-stream method with a combination of still frames and optical flow. Specifically, we fed the still RGB images into the pre-trained I3D models, which were originally trained for action recognition tasks. This allowed us to derive relevant spatio-temporal features from the video data.

\subsection{Sound} In the COGNIMUSE dataset, the accompanying sound for the movies can be classified as either speech or music. As part of our preprocessing method, each training and test sample is separated into non-overlapping fixed-length sound waveform files of approximately five seconds. This approach is similar to the one used in the work of ~\cite{thao2019multimodal}. Next, we extract features from the raw audio waveforms using SoundNet~\cite{aytar2016soundnet}.

\subsection{Text} In this paper, we leverage the multimodal properties of sound, video, and text as sources of supervision to predict emotion labels. By incorporating information from several modalities, we intend to increase the accuracy and robustness of emotion prediction. We use three modalities that are able to mutually supervise each other during the training process. Large-scale language models, such as BERT (Bidirectional Encoder Representations from Transformers)~\cite{devlin2018bert}, have shown significant performance in several natural language processing (NLP) tasks, showcasing their efficiency at both the word and sentence levels. These models have significantly advanced the field by capturing rich contextual information and semantic representations, allowing them to earn state-of-the-art results in diverse NLP applications. To extract bidirectional linguistic semantic data from the subtitles of each movie, we employ the power of the BERT (Bidirectional Encoder Representations from Transformers) model.Before using word2vec to embed each word from the subtitles, we apply a pre-processing step to remove English stop words ~\cite{mikolov2013distributed}.In our paper, the BERT model is employed, which enables us to understand the relationships between words, the meaning of sentences, and the overall structure of the subtitle text, which in turn contributes to our multimodal emotion anticipation model.

\subsection{Modality integration} To capture the correlations between diverse modalities and their influence on elicited emotion, we use the synchronous nature of sound, text, and visual input.
We utilize pairs of video and text (from the subtitles), as well as pairs of images and sound (from videos). Our proposed multimodal network, depicted in Figure ~\ref{fig:network}, leverages fully connected layers to incorporate and process diverse modalities for emotion classification. During training, the weights of the fully connected layers in our network are learned and optimized to precisely predict emotions. Subsequently, the outputs of these fully connected layers are concatenated and passed through another two fully connected layers. Using our multimodal network, we can predict emotion arousal and valence labels while individuals watch movies.

\section{Experiment}

To determine the performance of our model, we conduct tests using the COGNIMUSE dataset~\cite{zlatintsi2017cognimuse}, which
is composed of seven half-hour continuous movie clips. This dataset provides emotion annotations in terms of continuous arousal and valence ratings, ranging from -1 to 1. The primary focus of this work is to predict intended, expected and experienced emotions.In other words, we aim to understand and predict the emotions that individuals intend to convey, expect to feel, and actually experience while interacting with various multimedia content.

\subsection{Experimental setup} To ensure a comprehensive comparison, we also consider the results of intended emotions, which reflect the intentions of the filmmakers. These intentions are assessed by the same specialist who evaluates the other emotion categories. Additionally, to address potential challenges in emotion categorization, we employ seven different bins for each set of emotions. This method allows for a more nuanced analysis and interpretation of the emotions expressed in the multimedia content.
To evaluate the performance of our proposed models, we utilized a leave-one-out cross-validation method. This means that for each iteration, we train the model on all samples except one, and then test its performance on the left-out sample. This process is repeated for each sample in the dataset. To evaluate the accuracy of our emotion classification, we use two metrics, namely accuracy and accuracy±1. Accuracy measures the percentage of correctly classified emotions, while accuracy±1 considers emotions that are classified within one bin of the ground truth emotion.
In order to pre-process the data and remove noise, we apply Malandrakis' method~\cite{malandrakis2011supervised}. After noise removal, we further apply the Savitzky-Golay filter~\cite{savitzky1964smoothing} to smooth the data and decrease any remaining fluctuations. To guarantee consistency and comparability, we also rescale the preprocessed data within the range of -1 to 1.
The valence and arousal are independently classified into seven distinct classes by the models. To train these models, we utilize stochastic gradient descent (SGD) optimization algorithm. The models are trained with a learning rate of 0.005, weight decay of 0.005, and the softmax function with a temperature of T = 2.
During training, we run the models for 50 epochs with a batch size of 128. To avoid overfitting and enhance efficiency, we utilize early stopping with a patience of 25 epochs. 

\subsection{Results}

The results presented in table~\ref{table:results101} and \ref{table:results102} showcase the accuracy of our proposed models for the 7-class classification of intended emotion prediction and experienced emotion prediction, respectively. We followed the test-training split and evaluation design presented in \cite{thao2019multimodal} to ensure consistency in our evaluation methodology. Each video clip in our dataset is divided into five-second segments, with synchronized sound and subtitles for each segment. The raw RGB video frames are then fed into the ResNet-152, a pre-trained model on the ImageNet dataset, to extract features from each frame. These features are successively used as input to the final classification layer for emotion prediction.A completely connected layer is added on top of the retrieved features after applying max-pool to minimize dimensions, as seen in Fig~\ref{fig:network}. We employ the same methodology to extract scene features from locations. I3D models are used to extract motion features from snippet frames, and motion features are then layered with average pooling dimensional reduction before a fully connected layer is applied. The BERT model and SoundNet are utilized in a similar manner as the I3D model for extracting features from text and sound modalities, respectively. In our study, we investigate how diverse modalities, such as visual, sound, and text, contribute to the classification of viewers' emotional states. We use the network with fully connected layers for this analysis. In our study, we maintain a consistent architecture throughout the experiments, while changing the input features based on the specific modality being considered. We inspect diverse combinations of modalities, including audio, text, RGB frame features, Places features, and I3D features. Each modality is treated as a separate input to the architecture, allowing us to examine their individual contributions to the overall emotion recognition task. By systematically evaluating the performance of the model with different input features, we gain insights into the relative importance and effectiveness of each modality in capturing and predicting emotions. As shown in Table~\ref{table:results101}, text features provide a classification accuracy for arousal emotion labels that is higher than that of other modalities (image and motion). Sound modality gains high accuracy in valence emotion labels on experienced emotion annotation. According to the results presented in Table~\ref{table:results102},text modality obtains higher accuracy when predicting intended arousal emotion label, and the sound modality provides better accuracy in predicting intended valence emotion label.
The observed results, where text features perform well in anticipating intended arousal emotion labels and audio features perform better in predicting intended valence emotion labels, suggest that the influence of different modalities on emotions may vary. It is possible that text features, such as the semantic content derived from subtitles, play a more important role in conveying information related to arousal emotions. On the other hand, the audio features used, which capture the acoustic characteristics of the sound, may be more effective in capturing signals related to valence emotions.
We compare our approach with the state-of-the-art of ~\cite{malandrakis2011supervised, thao2019multimodal} in emotion prediction in Table ~\ref{table:results101},~\ref{table:results102}.

\section{Conclusion}
In this paper, we propose a novel deep convolutional network that leverages multimodal inputs, including sound, vision, and text, to learn representations for emotion identification.
Emotion classification is evaluated using both experienced and intended emotion annotations from the extended COGNIMUSE dataset. We train multiple model components and evaluate their performance when using diverse input modalities and their combinations.
The results of our experiments demonstrate significant improvements in both experienced and intended emotion annotations from the extended COGNIMUSE dataset, highlighting the efficacy of our approach in enhancing emotion recognition using multimodal information.


\bibliographystyle{IEEEbib}
\bibliography{strings,refs}

\end{document}